\documentclass{bmvc2k}
\usepackage[utf8]{inputenc}
\usepackage{multirow}
\usepackage{floatrow}
\usepackage{pgfplots}
\usepgfplotslibrary{groupplots}
\pgfplotsset{compat = 1.14}

\title{CAKE: Compact and Accurate K-dimensional representation of Emotion}
\addauthor{Corentin Kervadec\textsuperscript{*}}{corentin.kervadec@orange.com}{1}
\addauthor{Valentin Vielzeuf\textsuperscript{*}}{valentin.vielzeuf@orange.com}{12}
\addauthor{Stéphane Pateux}{stephane.pateux@orange.com}{1}
\addauthor{Alexis Lechervy}{alexis.lechervy@unicaen.fr}{2}
\addauthor{Frédéric Jurie}{frederic.jurie@unicaen.fr}{2}

\addinstitution{
Orange Labs, \\
Cesson-Sévigné, France
}
\addinstitution{
 Normandie Univ., UNICAEN, ENSICAEN, CNRS\\
 Caen, France
}

\runninghead{Kervadec, Vielzeuf, Pateux, Lechervy, Jurie}{CAKE}

\newfloatcommand{capbtabbox}{table}[][\FBwidth]

\begin{document}
%\begin{NoHyper}
\maketitle
%\end{NoHyper}
\begin{abstract}
Numerous models describing the human emotional states have been built by the psychology community. Alongside, Deep Neural Networks (DNN) are reaching excellent performances and are becoming interesting features extraction tools in many computer vision tasks. 
Inspired by works from the psychology community, we first study the link between the compact two-dimensional representation of the emotion known as {\em arousal-valence},  and discrete emotion classes (e.g. anger, happiness, sadness, \textit{etc.}) used in the computer vision community. It enables to assess the benefits -- in terms of discrete emotion inference -- of adding an extra dimension to arousal-valence (usually named dominance). Building on these observations, we propose CAKE, a 3-dimensional representation of emotion learned in a multi-domain fashion, achieving accurate emotion recognition on several public datasets. Moreover, we visualize how emotions boundaries are organized inside DNN representations and show that DNNs are implicitly learning arousal-valence-like descriptions of emotions. Finally, we use the CAKE representation to compare the quality of the annotations of different public datasets.
\end{abstract}
\section{Introduction}
\begin{figure}
\begin{floatrow}
\ffigbox[0.45\textwidth]{\includegraphics[width=\linewidth]{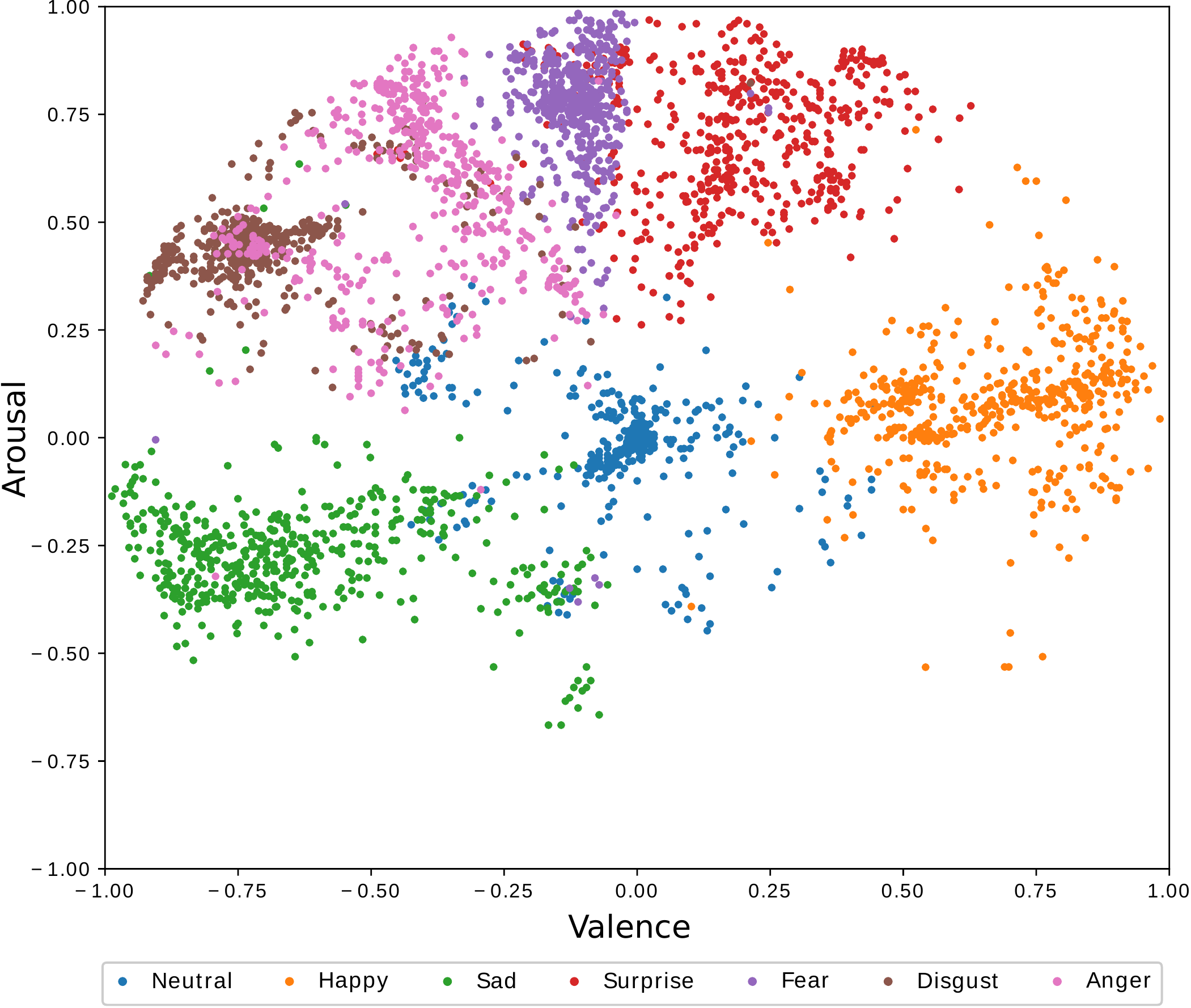}}{\caption{Comparison of the discrete and continuous (arousal-valence) representations using AffectNet's annotations~\cite{mollahosseini_affectnet:_2017}.}\label{fig:affecnet_arouval}}
\ffigbox[.45\textwidth]{\begin{tikzpicture}[scale=0.7]
\pgfplotsset{every tick label/.append style={font=\tiny}}
\begin{groupplot}[
group style={
    group name=my plots,
    group size= 1 by 1,
    xlabels at=edge bottom,
    ylabels at=edge left,
    },legend style={at={(0.7,0.2)},anchor=north, legend columns=-1}]

\nextgroupplot[xmode=log, mark size=2pt,ylabel=AffectNet Validation Accuracy (\%), xlabel=Representation Size, xtick={2,3,4,512}, ymax=87.1,log ticks with fixed point,tick style={grid=major}]
\addplot [color=brown, solid, mark=*] coordinates {
(2,83.0)
(3,86.0) 
(4,86.4) 
(512,86.6)
};
\end{groupplot}
\end{tikzpicture}}{\caption{Influence of adding supplementary dimensions to arousal-valence when predicting emotion on AffectNet~\cite{mollahosseini_affectnet:_2017}.}\label{fig:courbeAVdim}}
\end{floatrow}
\end{figure}

Facial expression is one of the most used human means of communication after language. Thus, the automated recognition of facial expressions -- such as emotions -- has a key role in affective computing, and its development could benefit human-machine interactions. 

Different models are used to represent human emotion states. Ekman~\textit{et al.}~\cite{ekman_constants_1971} propose to classify the human facial expression resulting from an emotion into six classes (\textit{resp.} happiness, sadness, anger, disgust, surprise and fear) supposed to be independent across the cultures. This model has the benefit of simplicity but could be not sufficient to address the whole complexity of human affect. Moreover it suffers from serious intra-class variations as, for instance, soft smile and laughing equally belong to \textit{happiness}. That is why Ekman's emotion classes are sometimes assembled into compound emotions~\cite{du2014compound} (\textit{e.g.} happily surprised).
Others have chosen to represent emotion with an n-dimensional continuous space, as opposite to the Ekman's discrete classes. Russel has built the \textit{Circumplex Model of Affect}~\cite{russell_circumplex_1980} in which emotion states are described by two values: arousal and valence.
\textit{Arousal} represents the excitation rate -- the higher the arousal is, the more intense the emotion is -- and \textit{valence} defines whether the emotion has a positive or a negative impact on the subject. Russels suggests in \cite{russell_circumplex_1980} that all Ekman's emotions~\cite{ekman_constants_1971} and compound emotions could be mapped in the \textit{circumplex model of affect}. Furthermore, this two-dimensional approach allows a more accurate specification of the emotional state, especially by taking its intensity into account.

A third dimension has been added by Mehrabian~\textit{et al.}~\cite{mehrabian1996pleasure} -- the \textit{dominance} -- which depends on the degree of control exerted by a stimulus. Last, Ekman and Friesen~\cite{ekman_measuring_1976} have come up with the \textit{Facial Action Code System} (FACS) using anatomically based action units. Developed for measuring facial movements, FACS is well suited for classifying facial expressions resulting from an affect. 

Based on these emotion representations, several large databases of face images have been collected and annotated according to emotion. EmotioNet~\cite{benitez-quiroz_emotionet:_2016} gathers faces annotated with Action Units~\cite{ekman_measuring_1976}; SFEW~\cite{dhall_static_2011}, FER-13~\cite{goodfellow_challenges_2013} and RAF~\cite{li_reliable_2017} propose images in the wild annotated in basic emotions; AffecNet~\cite{mollahosseini_affectnet:_2017} is a database annotated in both discrete emotion~\cite{ekman_constants_1971} and arousal-valence~\cite{russell_circumplex_1980}.

The emergence of these large databases has allowed to develop automatic emotion recognition systems, such as the recent approaches based on Deep Neural Networks (DNN). AffectNet's authors~\cite{mollahosseini_affectnet:_2017} use three AlexNet~\cite{krizhevsky_imagenet_2017} to learn respectively emotion classes, arousal and valence. In \cite{ng_deep_2015}, the authors make use of transfer learning to counteract the smallness of the SFEW~\cite{dhall_static_2011} dataset, by pre-training their model on ImageNet~\cite{deng_imagenet:_2009} and FER~\cite{goodfellow_challenges_2013}. In \cite{acharya_covariance_2018} authors implement {\em Covariance Pooling} using second order statistics when training on emotion recognition (on RAF~\cite{li_reliable_2017} and SFEW~\cite{dhall_static_2011}). 

Emotion labels, FACS and continuous representations have their own benefits -- simplicity of the emotion classes, accuracy of the arousal-valence, objectivity of the FACS, \textit{etc.} -- but also their own drawbacks --  imprecision, complexity, ambiguity, \textit{etc}. 
Therefore several authors have tried to leverage the benefits of all these representations. Khorrami \textit{et al.}~\cite{khorrami_deep_2015} first showed that neural networks trained for expression recognition implicitly learn facial action units.
Contributing to highlighting the close relation between emotion and Action Units, Pons \textit{et al.}~\cite{pons_multi-task_2018} learned a multitask and multi-domain ResNet~\cite{he_deep_2015} on both discrete emotion classes (SFEW~\cite{dhall_static_2011}) and Action Units (EmotioNet~\cite{benitez-quiroz_emotionet:_2016}). 
Finally, Li \textit{et al.}~\cite{li_reliable_2017} proposed a "\textit{Deep Locality-Preserving Learning}" to handle the variability inside an emotion class, by making classes as compact as possible. 

In this context, this paper focuses on the links between arousal-valence and discrete emotion representations for image-based emotion recognition. More specifically, the paper proposes a methodology for learning very compact embedding, with not more than 3 dimensions, performing very well on emotion classification task, making the visualization of emotions easy, and bearing similarity with the arousal-valence representation.

\section{Learning Very Compact Emotion Embeddings}
\label{methods}
\subsection{Some Intuitions About Emotion Representations}
\label{subsec:preliminary_study}

We first want to experimentally measure the dependence between emotion and arousal-valence as yielded in~\cite{russell_circumplex_1980}. We thus display each sample of the AffectNet~\cite{mollahosseini_affectnet:_2017} validation subset in the arousal-valence space and color them according to their emotion label (Figure~\ref{fig:affecnet_arouval}). For instance, a face image labelled as \textit{neutral} with an arousal and a valence of zero is located at the center of Figure~\ref{fig:affecnet_arouval} and colored in blue. It clearly appears that a strong dependence exists between discrete emotion classes and arousal-valence. Obviously, it is due in part to the annotations of the AffectNet~\cite{mollahosseini_affectnet:_2017} dataset, as the arousal-valence have been constrained to lie in a predefined confidence area based on the emotion annotation. Nevertheless, this dependence agrees with the \textit{Circumplex Model of Affect}~\cite{russell_circumplex_1980}.

To evaluate further how arousal-valence representation is linked to emotion labels, we train a classifier made of one  fully connected layer\footnote{By "fully connected layer" we denote a linear layer with biases and without activation function.} (fc-layer) to infer emotion classes from arousal-valence values provided by AffectNet~\cite{mollahosseini_affectnet:_2017} dataset. We obtain the accuracy of 83\%, confirming that arousal-valence can be an excellent \textit{2-d} compact emotion representation.  

This raises the question of the optimality of this 2-\textit{d} representation. Would adding a third dimension to arousal-valence make the classification performance better? To address this question, we used the 512-\textit{d} hidden representation of a ResNet-18~\cite{he_deep_2015} trained to predict  discrete emotions on the AffectNet dataset~\cite{mollahosseini_affectnet:_2017}. This representation is then projected into a more compact space using a fc-layer outputting $k$ dimensions, which are concatenated with the arousal-valence values. On top of this representation, we add another fc-layer predicting emotion classes. The two fc-layers are finally trained using Adam optimizer~\cite{kingma2014adam}.
Adding 1 dimension to arousal-valence gives a gain of +3 points on the accuracy. It agrees with the assumption that a three-dimensional representation is more meaningful than a two-dimensional one~\cite{mehrabian1996pleasure}. The benefit of adding more than 1 dimension is exponentially decreasing; with +512 dimensions, the gain is only of +0.6 points compared to adding 1 dimension, as shown in Figure~\ref{fig:courbeAVdim}.

From these observations, the use of a compact representation seems to be consistent with discrete emotion classes, as it enables an accuracy of 83\% and 86\% -- respectively for a 2-\textit{d} and a 3-\textit{d} representation -- and it even may allow to describe affect states with more contrast and accuracy. 
Even if arousal-valence is a good representation for emotion recognition, the question of its optimality has not been answered by these preliminary experiments. In other words, is it possible to learn 2-\textit{d} (or 3-\textit{d}) embedding better than those built on arousal-valence? We positively answer this question in Section~\ref{subsec:method}. 

\subsection{Learning Compact and Accurate Representations of Emotions}
\label{subsec:method}
\begin{figure}
    \centering
    \includegraphics[width=\linewidth]{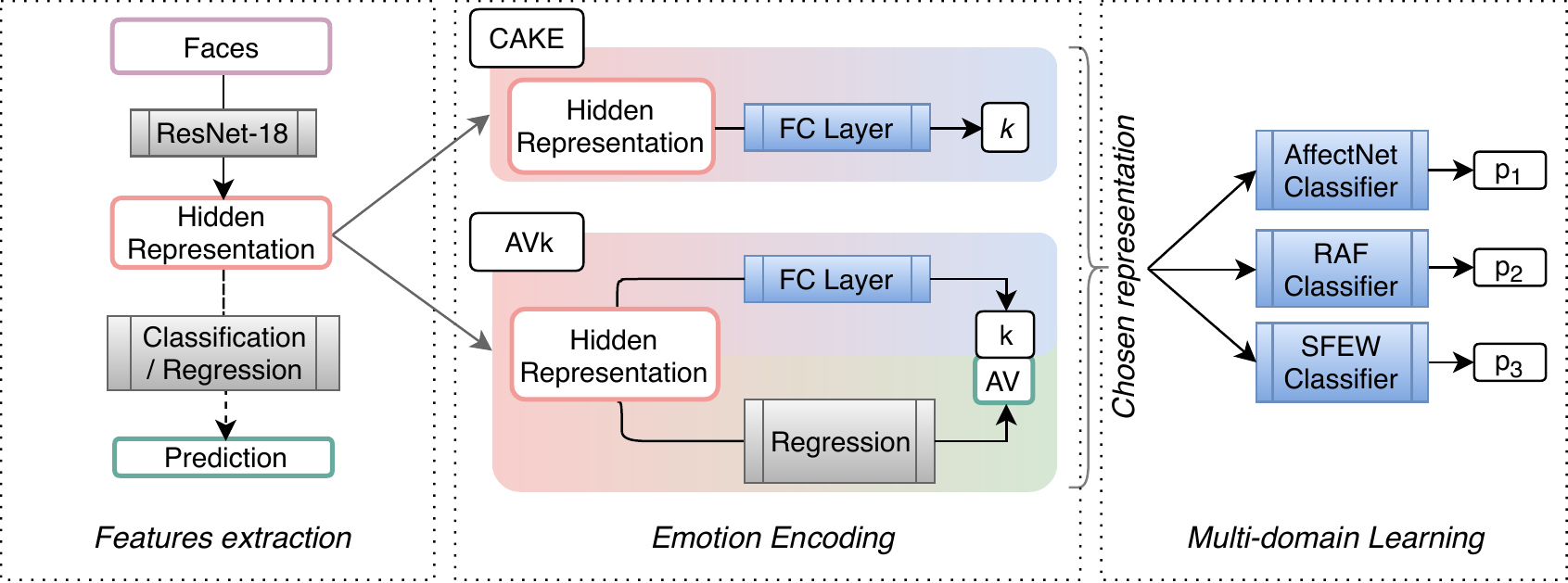}
    \caption{Our approach's overview. Left: we use a ResNet-18 previously trained for discrete emotion recognition or arousal valence regression to extract 512-d hidden representations from face images. Center: using these hidden representations, CAKE or AVk representations (center) are learned to predict discrete emotions. Right: the learning process is multi-domain, predicting emotions on three different datasets with three different classifiers. Gray blocks are non-trainable weights while blue blocks are optimized weights.}
    \label{fig:cross_arch}
\end{figure}

Based on the previous observations, this section proposes a methodology for learning a compact embedding for emotion recognition from images.
\paragraph{Features extraction}
The basic input of our model is an image containing one face displaying a given emotion. We first extract 512-\textit{d} features specialized in emotion recognition.
So as to, we detect the face, align its landmarks by applying an affine transform and crop the face region. The so-obtained face is then resized into $224\times224$ and fed to a ResNet-18~\cite{he_deep_2015} network (Figure~\ref{fig:cross_arch}, \textit{Features extraction}). The face image is augmented (\textit{e.g.} jittering, rotation), mostly to take the face detector noise into account. We also use cutout~\cite{devries_improved_2017} -- consisting in randomly cutting a $45\times45$ pixels sized patch from the image -- to regularize and improve the robustness of our model to facial occlusions.
Our ResNet outputs 512-\textit{d} features, on top of which a fc-layer can be added. At training time, we also use dropout~\cite{srivastava2014dropout} regularization.
The neural network can be learned from scratch on two given tasks: discrete emotion classification or arousal-valence regression. 

\paragraph{Compact emotion encoding}
Compact embedding is obtained by projecting the 512-\textit{d} features provided by the ResNet-18 (pretrained on discrete emotion recognition) into smaller k-dimensional spaces (Figure~\ref{fig:cross_arch}, \textit{Emotion Encoding}) in which the final classification is done.
The $k$ features may be seen as a compact representation of the emotion, and the performance of the classifier can be measured for different values of $k$. CAKE-2, CAKE-3, \textit{etc.}, denote such classifiers with $k=2$, $k=3$, \textit{etc}.

In the same fashion we can train the ResNet-18 using arousal-valence regression. In this case, the so-obtained arousal-valence regressor can be used to infer arousal-valence values from novel images and concatenate them to the $k$ features of the embedding. Thus we reproduce here the exact experiment done in Section~\ref{subsec:preliminary_study} in order to assess the benefit of a third (or more) dimension. The difference is that arousal-valence are not ground truth values but predicted ones. These methods are denoted as AV1, AV2, AV3, \textit{etc.} for the different values of $k$.

\paragraph{Domain independent embedding}
As we want to ensure a generic compact enough representation, independent of the datasets, we learn the previously described model jointly on several datasets, without any further fine-tuning.

Our corpus is composed of AffectNet~\cite{mollahosseini_affectnet:_2017}, RAF~\cite{li_reliable_2017} and SFEW~\cite{dhall_static_2011}, labelled with seven discrete emotion classes: \textit{neutral}, \textit{happiness}, \textit{sad}, \textit{surprise}, \textit{fear}, \textit{disgust} and \textit{anger}. %We split our corpus into two subsets, one for training and the other for testing. 
Our training subset is composed of those of AffectNet (283901 elts., \textit{95.9\%} of total), RAF (11271 elts., \textit{3.81\%} of total) and SFEW (871 elts., \textit{0.29\%} of total). Our testing subset is composed of the subsets commonly used for evaluation in the literature (\textit{validation} of SFEW and AffecNet, \textit{test} of RAF).

To ease the multi-domain training, we first pre-train our features extractor model on AffectNet and freeze its weights. Then we apply the same architectures as described before, but duplicate the last fc-layer in charge of emotion classification in three dataset-specific layers (Figure~\ref{fig:cross_arch}, \textit{multi-domain learning}). The whole model loss is a modified softmax cross entropy defined as follows:
\begin{equation}
Loss=\frac{1}{N} \sum_{i=1}^{N} \sum_{j=1}^{3} w_{class}^{i,j} w_{dataset}^{j} \ E(y^i,\hat{y}^{i,j})
\end{equation}
where $j$ is ranging in [AffectNet, RAF, SFEW],  $y^i$ is the label of $i^{th}$ element, $\hat{y}^{i,j}$ is the prediction of the $j^{th}$ classifier on the $i^{th}$ element, E is the softmax cross entropy loss,  $N$ is the number of elements in the batch, $w_{class}^i$ is a weight given to the $i^{th}$ element of the batch depending on its emotion class and $w_{dataset}^{j}$ is a weight given to the $j^{th}$ classifier prediction.
Each sample of the multi-domain dataset is identified according to its original database, allowing to choose the correct classifier's output when computing the softmax cross entropy.

The $ w_{class}$ weight is defined as:
$
    w_{class}^{i,j} = \frac{N_{total}^j}{N_{class}^{i,j} \times nbclass}
$
where $N_{total}^j$ is the number of elements in the $j^{th}$ dataset, $N_{class}^{i,j}$ is the number of elements in the class of the $i^{th}$ element of the $j^{th}$ dataset and $nbclass$ is the number of classes (7 in our case). The goal here is to fix the important class imbalance in the dataset by forcing to fit the uniform distribution, as previously done by \cite{mollahosseini_affectnet:_2017}.

The $w_{dataset}$ weight permits to take the imbalance between dataset's sizes into account.
\begin{align}
 w_{dataset}^{j}= \left\{
 \begin{array}{cl}
      \frac{1}{\log N_{total}^{j}}  & sample \in j^{th} dataset\\
      0       & sample \notin j^{th} dataset
\end{array}
\right.
\end{align}

We thus define a global loss enabling to optimize the last two layers of our model (namely \textit{Emotion Encoding} and \textit{Multi-domain Learning}  in Figure~\ref{fig:cross_arch}) on the three datasets at the same time. The dimension $k$ (or $k+2$ in the case of the arousal-valence approach) can easily be changed and help to evaluate the interest of supplementary dimensions for emotion representation. 
\section{Experiments}
\label{results}
\subsection{Evaluation Metrics}
We measure the classification performance with the \textit{accuracy} and the \textit{\textbf{macro} F1 Score}~\eqref{eq:f1}. \textit{Accuracy} measures the number of correctly classified samples. Instead of accuracy, we prefer \textit{macro F1 score} which gives the same importance to each class:
%\begin{equation}accuracy=\frac{1}{N_{elts}}\sum_{i \in C}tp_i
%\label{eq:accuracy}
%\end{equation}
\begin{equation}\begin{aligned}
F_{1macro}=\frac{1}{N_c}\sum_{i}^{N_c}F_{1i} \quad
F_{1i}=2\frac{prec_i \cdot rec_i}{prec_i+rec_i} \quad
prec_i=\frac{tp_i}{tp_i+fp_i} \quad
rec_i=\frac{tp_i}{tp_i+fn_i} \quad
\label{eq:f1}
\end{aligned}
\end{equation}
where $i$ is the class index; $prec_i$, $rec_i$  and $F_{1i}$ are the precision, the recall and the F1-score of class $i$; $N_c$ is the number of classes; $tp$, $fp$ and $fn$ are the true positives, false positives and false negatives rates. All scores are averaged on 10 runs, with different initializations, and given with associated standard deviations, on our multi-domain testing subset.

\subsection{Compactness of the Representation}
\label{subsec:compactness}
\begin{figure}
\begin{floatrow}
\ffigbox[.5\textwidth]{\begin{tikzpicture}[scale=0.75]
\pgfplotsset{every tick label/.append style={font=\tiny}}
\begin{groupplot}[
group style={
    group name=my plots,
    group size= 1 by 1,
    xlabels at=edge bottom,
    ylabels at=edge left,
    },legend style={at={(0.7,0.2)},anchor=north, legend columns=-1}]

\nextgroupplot[xmode=log, mark size=2pt,ylabel=Multi-Domain Test F1 Score (\%), xlabel=Representation Size (log scale), xtick={2,3,4,5,6,7,512}, ymax=64,log ticks with fixed point,tick style={grid=major}]
\addplot [color=black, dashdotted, mark=otimes*] coordinates {
(2, 54.37)
(3, 59.39)
(4, 61.07)
(5, 62.08)
(6, 62.15)
(7, 62.50)
(512, 62.64)
};
\addplot [color=brown, solid, mark=x] coordinates {
(2, 56.79)
(3, 57.28)
(4, 58.60)
(5, 59.29)
(6, 59.50)
(7, 59.72)
(512, 61.72)
};
\legend{CAKE, AVk}

%\nextgroupplot[xmode=log, mark size=2pt,ylabel=AffectNet Validation Accuracy (\%), xlabel=Representation Size, xtick={2,3,5,7,512},log ticks with fixed point]
%\addplot [color=brown, solid, mark=*] coordinates {
%(2,83.0)
%(3,86.0) 
%(4,86.4) 
%(512,86.6)
%};
%\legend{Arousal-Valence Labels Based Representation}

\end{groupplot}

\end{tikzpicture}
}{\caption{Influence of representation size on the multi-domain F1 score.}\label{fig:repsize}}

\capbtabbox{%\begin{tabular}{|l|l|l|l|}

\begin{tabular}{|l|l|l|l|}
\hline
Dataset                                                                & \multicolumn{2}{l|}{Rep. \& Dim.} & F1 Score                \\ \hline
\multirow{3}{*}{\begin{tabular}[c]{@{}l@{}}Affect\\ -Net\end{tabular}} & CAKE-3            & 3            & \textbf{58.1 $\pm$ 0.5} \\
                                                                       & AV1               & 3            & 55.6 $\pm$ 0.5          \\
                                                                       & AV                & 2            & 55.8 $\pm$ 0.0          \\
                                                                       & CAKE-2            & 2            & 52.1 $\pm$ 0.4          \\ \hline
\multirow{3}{*}{SFEW}                                                  & CAKE-3            & 3            & \textbf{34,1 $\pm$ 1.0} \\
                                                                       & AV1               & 3            & 30.2 $\pm$ 0.8          \\
                                                                       & AV                & 2            & 33.3 $\pm$ 0.1          \\
                                                                       & CAKE-2            & 2            & 28.0 $\pm$ 0.8          \\ \hline
\multirow{3}{*}{RAF}                                                   & CAKE-3            & 3            & \textbf{64.4 $\pm$ 0.5} \\
                                                                       & AV1               & 3            & 63.0 $\pm$ 0.9          \\
                                                                       & AV                & 2            & 61.2 $\pm$ 0.2          \\
                                                                       & CAKE-2            & 2            & 60.6 $\pm$ 1.9          \\ \hline
\end{tabular}

}{\caption{Evaluation of compact representations on AffectNet, SFEW, RAF.}
\label{table:compact}}
\end{floatrow}
\end{figure}

We first evaluate the quality of the representations in a multi-domain setting. Table~\ref{table:compact} reports the  F1-score of CAKE-2, AV, CAKE-3 and AV1 trained on three datasets with three different classifiers, each one being specialized on a dataset as explained in Section~\ref{methods}. Among the 2-\textit{d} models (AV and CAKE-2), AV is better, taking benefits from the knowledge transferred from the AffectNet dataset. This is not true anymore for the 3D models, where CAKE-3 is better than AV1, probably because of its greater number of trainable parameters.

To validate the hypothesis of the important gain brought by adding a third dimension, we run the "CAKE" and "AVk" experiments with different representation sizes. To simplify the analysis of the results, we plot in Figure~\ref{fig:repsize} a multi-domain F1-score, \textit{i.e.} the weighted average of the F1-scores according to the respective validation set sizes.
We observe that the gain in multi-domain F1-score is exponentially decreasing for both representations -- note that the representation size axis is in log scale -- and thus the performance gap between a representation of size 2 and size 3 is the more important. We also observe that "CAKE" representations still seem to yield better results than "AVk" when the representation size is greater than 2. 

These first experiment shows that a very compact representation can yield good performances for emotion recognition. It also is in line with the "dominance" dimension hypothesis, as a third dimension brought the more significant gain in performance. After 3 dimensions, the gain is much less significant.

\subsection{Accuracy of the Representation}
To evaluate the efficiency of the CAKE-3 compact representation, we compare its accuracy with 
state-of-the-art approaches (Table~\ref{table:comparison}) on the public datasets commonly used in the literature for evaluation (\textit{validation} of SFEW and AffecNet, \textit{test} of RAF). In order to get a fair comparison, we add a \textit{"Rep. Dim."} column corresponding to the size of the last hidden representation -- concretely, we take the penultimate fully connected output size.
We report the scores under the literature's metrics, namely the mean of the per class recall for RAF~\cite{li_reliable_2017} and the accuracy for SFEW~\cite{dhall_static_2011} and AffectNet~\cite{mollahosseini_affectnet:_2017}. To the best of the author's knowledge no other model has been evaluated before on the AffectNet's seven classes.

CAKE-3 is outperformed by Covariance Pooling~\cite{acharya_covariance_2018} and Deep Locality Preserving~\cite{li_reliable_2017}.
Nevertheless, it is still competitive as the emotion representation is far more compact -- 3-\textit{d} \textit{versus} 2000-\textit{d} -- and learned in a multi-domain fashion. Moreover, we gain 1 point on RAF when we compare to models of same size (2 millions parameters), \textit{e.g.} \textit{Compact Model}~\cite{kuo2018compact}. These results support the conclusion made in \ref{subsec:compactness}, as we show that a compact representation of the emotion learned by small models is competitive with larger representations. This finally underlines that facial expressions may be encoded efficiently into a 3-\textit{d} vector and that using a large embedding on small datasets may lead to exploit biases of the dataset more than to learn emotion recognition.

\begin{table}
\begin{tabular}{c|c|ccc|}
\cline{2-5}
    & Rep. Dim. & \multicolumn{1}{c}{RAF~\cite{li_reliable_2017}} & \multicolumn{1}{c}{SFEW~\cite{dhall_static_2011}} & AffectNet~\cite{mollahosseini_affectnet:_2017}    \\ \hline
\multicolumn{1}{|c|}{\multirow{2}{*}{Covariance Pooling~\cite{acharya_covariance_2018}}}    & 2000                & 79.4                     & -                         & -              \\ \cline{2-5} 
\multicolumn{1}{|c|}{}                                                                      & 512                 & -                        & 58.1                     & -              \\ \hline
\multicolumn{1}{|c|}{Deep Locality Preserving~\cite{li_reliable_2017}}                      & 2000                & 74.2                    & 51.0                     & -              \\ \hline
%\multicolumn{1}{|c|}{Center Loss~\cite{wen2016discriminative}}                             & 2000                & 72.87                    & -                         & -              \\ \hline
\multicolumn{1}{|c|}{Compact Model~\cite{kuo2018compact}}                                   & 64                  & 67.6                    & -                         & -              \\ \hline
\multicolumn{1}{|c|}{VGG\cite{li_reliable_2017}}                                            & 2000                & 58.2                    & -                         & -              \\ \hline
\multicolumn{1}{|c|}{Transfer Learning~\cite{ng_deep_2015}}                                 & 4096                & -                        & 48.5                      & -              \\ \hline\hline
%\multicolumn{1}{|c|}{EmotiW 2015~\cite{kim2016hierarchical}}                                & 3072                & -                        & 52.5                      & -              \\ \hline\hline
\multicolumn{1}{|c|}{{ours (CAKE-3)}}                                                      & {3}          & {68.9}           & {44.7}            & {58.2} \\ \hline
\multicolumn{1}{|c|}{{ours (Baseline)}}                                                & {512}        & {71.7}           & {48.7}            & {61.7} \\ \hline
\end{tabular}
\caption{Accuracy of our model regarding state-of-the-art methods. The size of the representation is taken into account. Metrics are the average of per class recall for RAF and accuracy for SFEW and AffectNet.}
\label{table:comparison}
\end{table}

Our experiments also allow to perform a cross-database study as done in \cite{li_reliable_2017}. This study consists in evaluating a model trained on dataset B on a dataset A. Thereby we obtain Table~\ref{table:cross_f1} with the evaluation of each classifier on each dataset.
Results on SFEW~\cite{dhall_static_2011} -- trained or evaluated -- are constantly lower than others, with a higher standard deviation. This could be due to the insufficient number of samples in the SFEW training set or more probably to the possible ambiguity in the annotation of SFEW compared to AffectNet and RAF. Confirming this last hypothesis, the \textit{RAF classifier} has the better generalization among the datasets. It is in line with the claim of Li~\textit{et al.}~\cite{li_reliable_2017} that RAF has a really reliable annotation with a large consensus between different annotators. Finally, it also underlines the difficulty to find a reliable evaluation of an emotion recognition system because of the important differences between datasets annotations.

\begin{table}
\begin{tabular}{cl|l|l|l|}
\cline{3-5}
\multicolumn{1}{l}{}                                       &           & \multicolumn{3}{c|}{Dataset}                     \\ \cline{3-5} 
\multicolumn{1}{l}{}                                       &           & AffectNet         & SFEW              & RAF      \\ \hline
\multicolumn{1}{|c|}{\multirow{3}{*}{Classifier}} & AffectNet & \textbf{58.1 } \textit{($\pm$ 0.5)} & 27.6 \textit{($\pm$ 2.6)}          & 53.8  \textit{($\pm$ 0.6)} \\ \cline{2-5} 
\multicolumn{1}{|c|}{}                                     & SFEW      & 35.1  \textit{($\pm$ 2.1)}          & \textbf{34.1 } \textit{($\pm$ 1.0)} & 47.3  \textit{($\pm$ 1.2)} \\ \cline{2-5} 
\multicolumn{1}{|c|}{}                                     & RAF       & 51.8  \textit{($\pm$ 0.4)}          & 31.5  \textit{($\pm$ 1.7)}         & \textbf{64.4 } \textit{($\pm$ 0.6)} \\ \hline
\end{tabular}
\caption{Cross-database evaluation on CAKE-3 model (F1-Score).}
\label{table:cross_f1}
\end{table}
\subsection{Visualizing Emotion Maps}
Visualizations are essential to better appreciate how DNN performs classifications, as well as to visualize emotion boundaries and their variations across datasets. Our visualization method consists in densely sampling the compact representation space -- 2-\textit{d} or 3-\textit{d} -- into a mesh grid, and feeding it to a formerly trained model -- AV, CAKE-2 or CAKE-3 -- in order to compute a dense map of the predicted emotions. Not all the coordinates of the mesh grid belong to real emotions and some of them would never happen in real applications. 

The construction of the mesh grid depends on the model to be used. 
For the AV and the CAKE-2 models, we have simply built it using 2d vectors with all values ranging in intervals containing maximum and minimum values of the coordinates observed with real images. As the CAKE-3 model is dealing with a three-dimensional representation, it is not possible to visualize it directly on a plane figure.
To overcome this issue we modify CAKE-3 into a CAKE-3-Norm representation where all the coordinates are constrained to be on the surface of the unit sphere, and visualize spherical coordinates. 
Even if CAKE-3-Norm shows lower performances (about 2 points less than CAKE-3), the visualization is still interesting, bringing some incentives about what has really been learned.

\begin{figure}[hb]
    \begin{minipage}{.86\linewidth}
    \includegraphics[width=\linewidth]{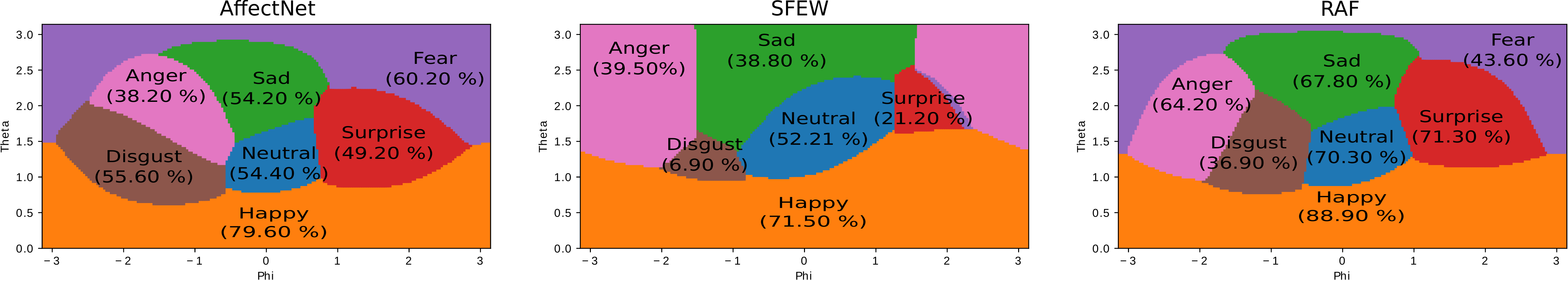}
    \end{minipage}
    \begin{minipage}{.86\linewidth}
    \includegraphics[width=\linewidth]{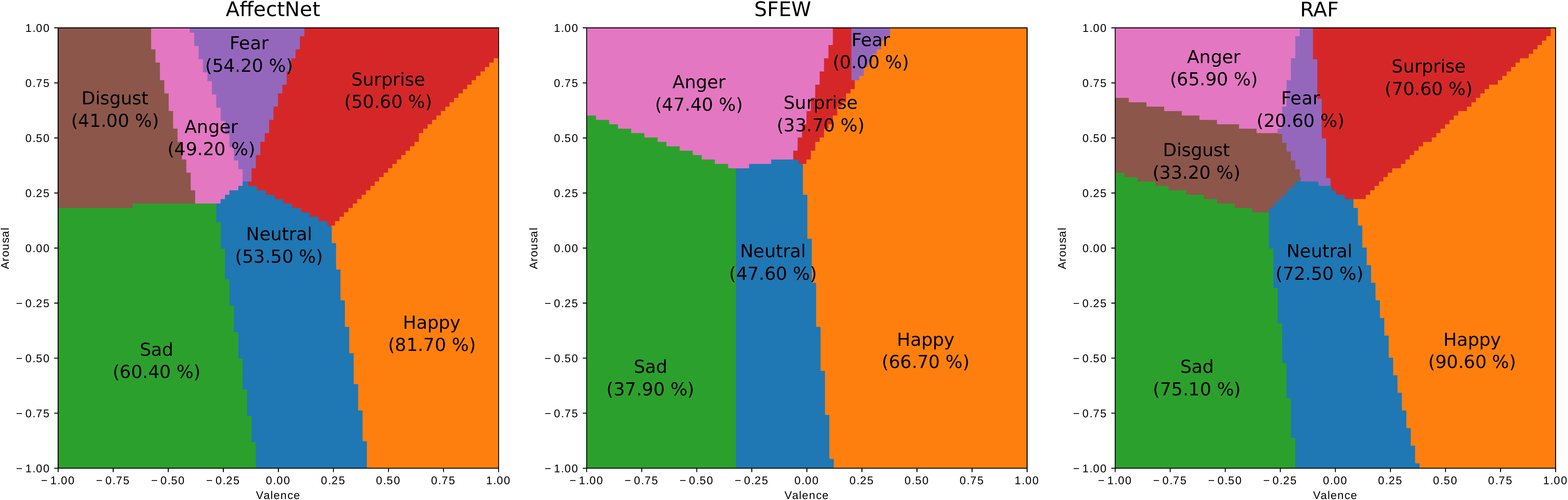}
    \end{minipage}
    \begin{minipage}{.86\linewidth}
    \includegraphics[width=\linewidth]{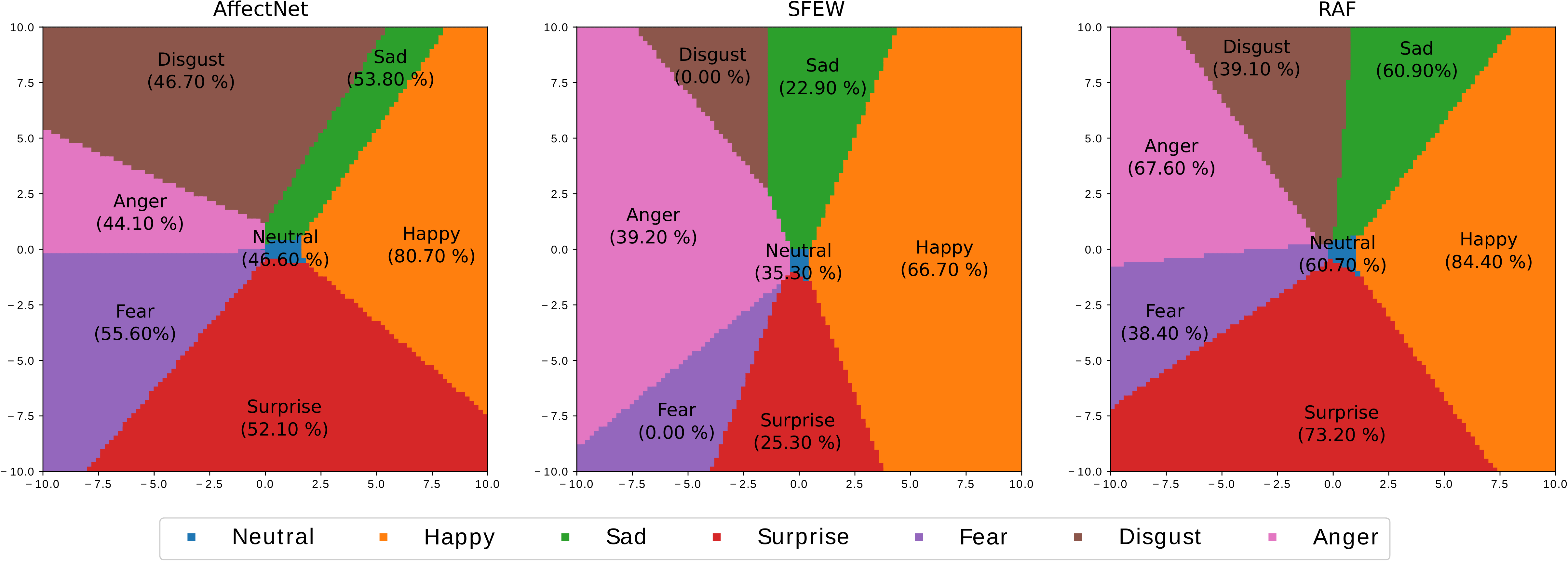}
    \end{minipage}
  \caption{Visualization of CAKE-3-Norm, AV and CAKE-2. Rows indicate evaluated representation -- \textit{resp.} from top to down: CAKE-3-Norm, AV, CAKE-2 -- and columns indicate datasets -- \textit{resp.} from left to right: AffectNet~\cite{mollahosseini_affectnet:_2017}, SFEW~\cite{dhall_static_2011} and RAF~\cite{li_reliable_2017}.}
  \label{fig:visu}
\end{figure}

Figure~\ref{fig:visu} shows the visualization results  for CAKE-3-Norm, AV and CAKE-2 representations (\textit{resp.} from top to down).
Each dot is located by the coordinates of its compact representation -- \((arousal, valence)\) for AV, \((k_1, k_2)\) for CAKE-2 and spherical coordinates ($\phi$ and $\theta$) for CAKE-3-Norm -- and colored according to the classifier prediction. The per class macro F1-score is displayed inside each emotion area.

First, each compact representation -- CAKE-2, CAKE-3-Norm and AV -- exhibits a strong consistency across the datasets (in Figure~\ref{fig:visu}, compare visualizations on the same row). Indeed, the three classifiers show a very similar organization of the emotion classes, which is demonstrating the reliability of the learned representation. Thereby, the \textit{neutral} class -- in blue -- is always placed at the origin and tends to neighbor all other classes. It is in line with the idea of neutral as an emotion with a very low intensity. Nevertheless, we can witness small inter-dataset variations, especially on SFEW~\cite{dhall_static_2011} (in Figure~\ref{fig:visu}, middle column) with \textit{disgust} and \textit{fear} -- \textit{resp.} brown and purple -- which are almost missing. This underlines the disparities of annotations across the datasets and confirms the need of multi-domain frameworks when wishing to achieve a more general emotion recognition model.

Second, we can analyze variations between the different representations for a given dataset (in Figure~\ref{fig:visu}, compare visualizations on the same column). As AV is based on arousal-valence, we observe the same emotion organization as in Figure \ref{fig:affecnet_arouval}. Especially, as the majority of the AffectNet's training (and validation) samples have a positive arousal, the classifier do not use the whole space (in Figure~\ref{fig:visu}, second row: see green, blue and orange areas) unlike CAKE-2 and CAKE-3 which are not constrained by arousal-valence. 

We can find many similarities between these three representations, but the most impressive come across when comparing CAKE-2 and AV. Despite the inequality of scaling -- which causes the \textit{neutral} area (blue) to be smaller in CAKE-2 -- AV and CAKE-2 compact representations are very close. Indeed, the area classes are organized exactly in the same fashion. The only difference is that for AV they are disposed in a clockwise order around \textit{neutral} whereas for CAKE-2 they are disposed in an anticlockwise order. This observation shows that a DNN trained on the emotion recognition classification is able to learn an arousal-valence-like representation of the emotion. It contributes -- along with Khorrami~\cite{khorrami_deep_2015} who points that DNNs trained to recognize emotions are learning action units~\cite{ekman_measuring_1976} -- to bring the dependence across the emotion representations in the forefront.

\section{Conclusion}
This work proposes a comprehensive analyze on how a DNN can describe emotional states. To this purpose, we first studied how many dimensions are sufficient to accurately represent an emotion resulting from a facial expression. We then conclude that three dimensions are a good trade-off between accuracy and compactness, agreeing with the arousal-valence-dominance~\cite{russell_circumplex_1980}\cite{mehrabian1996pleasure} psychologist model.
Thereby, we came up with a DNN providing a 3-dimensional compact representation of emotion, learned in a multi-domain fashion on RAF~\cite{li_reliable_2017}, SFEW~\cite{dhall_static_2011} and AffecNet~\cite{mollahosseini_affectnet:_2017}. We set up a comparison with the state-of-the-arts and showed that our model can compete with models having much larger feature sizes. It proves that bigger representations are not necessary for emotion recognition. In addition, we implemented a visualization process enabling to qualitatively evaluate the consistency of the compact features extracted from emotion faces by our model. We thus showed that DNN trained on emotion recognition are naturally learning an arousal-valence-like~\cite{russell_circumplex_1980} encoding of the emotion. As a future work we plan to also apply state-of-the-art techniques -- as Deep Locality Preserving Loss~\cite{li_reliable_2017} or Covariance Pooling~\cite{acharya_covariance_2018} -- to enhance our compact representation. In addition, nothing warranty that the learned CAKE bears the same semantic meanings as arousal-valence-dominance does: further interpreting the perceived semantic of the dimensions would therefore be an interesting piece of work.

%  In addition, we aim to search if our CAKE-3 representation 
\bibliography{egbib}
\end{document}